\documentclass[runningheads]{llncs}
\usepackage{graphicx}

\usepackage{tikz}
\usepackage{comment}
\usepackage{amsmath,amssymb} 
\usepackage{color}

\usepackage[accsupp]{axessibility}  

\begin{document}
\pagestyle{headings}
\mainmatter
\def\ECCVSubNumber{2130}  

\title{Space Time Recurrent Memory Network} 

\titlerunning{Space Time Recurrent Memory Network}
%
\author{Hung Nguyen \and
Chanho Kim \and
Fuxin Li}
\authorrunning{H. Nguyen \and C. Kim \and F. Li}
%
\institute{Oregon State University\\
\email{\{nguyehu5,kimchanh,Fuxin.Li\}@oregonstate.com}}
\maketitle

\begin{abstract}
Transformers have recently been popular for learning and inference in the spatial-temporal domain. However, their performance relies on storing and applying attention to the feature tensor of each frame in video. Hence, their space and time complexity increase linearly as the length of video grows, which could be very costly for long videos.
We propose a novel visual memory network architecture for the learning and inference problem in the spatial-temporal domain. We maintain a fixed set of memory slots in our memory network and propose an algorithm based on Gumbel-Softmax to learn an adaptive strategy to update this memory. Finally, this architecture is benchmarked on the video object segmentation (VOS) and video prediction problems. We demonstrate that our memory architecture achieves state-of-the-art results, outperforming transformer-based methods on VOS and other recent methods on video prediction while maintaining constant memory capacity independent of the sequence length.
\end{abstract}

\section{Introduction}

\label{sec:intro}

Network architectures for spatial-temporal reasoning are of fundamental importance to computer vision systems. Most real-life reasoning problems are presented in space-time, including but not limited to autonomous driving, video object segmentation, navigation, planning, etc. A great earlier example of a spatial-temporal reasoning model is the convolutional LSTM~\cite{shi2015convolutional} in which a convolutional network generates a spatially indexed memory tensor and recurrently updates the memory tensor over time as the network processes more input frames.

There are two main benefits to the recurrent memory model. First, it is \textbf{online}, meaning that the system can generate appropriate output at any given time of the sequence. The system would do so by taking into account the memory which has evolved through all the previous frames, hence potentially taking into account long-term information. Second, the memory size is \textbf{constant}, as the memory is updated at each time 
with the new information from each frame. Such properties make the recurrent memory model ideal for embodied systems that require real-time processing.

However, one drawback of the convolutional LSTM or other similar approaches  is that the memory may confuse itself over \textit{multi-modality}. Since there is only one feature tensor stored as memory, the system often has to face a dilemma when two equally valuable templates exist: they have to either only remember one of them or somehow combine them into one template. However, the combination may become more ambiguous  and lead to reduced performance when many diverse templates need to be remembered. Another drawback of such approaches is that they do not handle object motion well when strong information about moving objects in the scene. Although the memory could be indexed spatially, such indexing is rigid. As a result, when the object moves in space, the stored memory at the indexed location may not be accurate anymore.

More recently, spatio-temporal transformers (STM) \cite{STM} were  proposed as a different approach for spatio-temporal reasoning. Motivated by transformers in natural language processing, each of previous frames forms two spatially-indexed feature tensors called the query and the key in STM. Then, a new frame is also converted to the query and key tensors, which are used to match with the previous frames to compute attention weights. These weights are used in a weighted sum of the spatially-indexed \textit{value} tensors of the previous frames. STM has achieved state-of-the-art performance in Video Object Segmentation (VOS) which is one of the most important and difficult spatio-temporal reasoning problems. However, similarly with most transformer approaches, STM needs to compute and store tensors for each frame, leading to significantly increased time and memory cost for long videos. The walkaround in \cite{STM} and many subsequent work \cite{yang2021aot,cheng2021stcn} was to store one set of memory tensors for every 5th frame, but this would not extend easily to videos with thousands of frames. 

\begin{figure*}[ht]
\begin{center}
\includegraphics[width=0.9\linewidth]{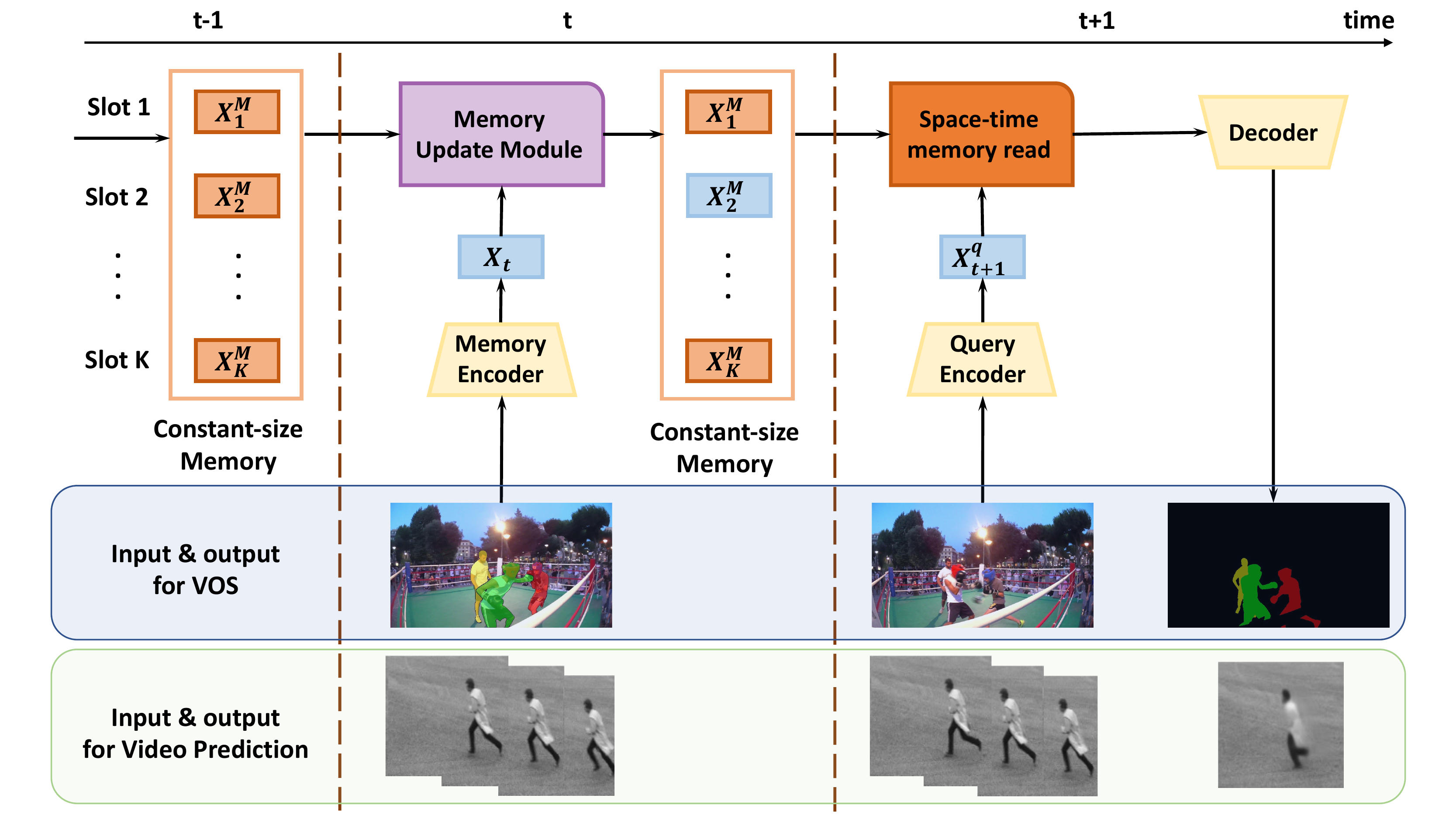}
\end{center}


\caption{\textbf{The overview of our framework}. We evaluate our framework on VOS and video prediction. Our framework consists of two stages. 1) Memory Update: In this stage, the memory encoder generates a new feature map $X_t$ by taking the current frame and its corresponding object masks as input in the case of VOS (taking 3 consecutive frames as input in the case of video prediction). This feature map is then used to update the memory (Sec. \ref{sec: memory update}). 2) Inference stage: The query encoder generates a feature map $X_{t+1}^q$ by taking the query image as input in the case of VOS (taking three consecutive query frames as input in the case of video prediction). The space-time memory read module takes the feature map and current memory as input and retrieves information relevant to the target task as another feature map (Sec. \ref{sec: inference stage}). This output feature map from the memory read module is finally decoded into our predicted mask (or the next frame in the case of video prediction) which will be then used to update the memory again. (Best viewed in color)}\label{fig:overall framework}

\end{figure*}

In this paper, we propose a novel memory network, the Space-Time Recurrent Memory Network (STReMN, pronounced as ``strem'') which attempts to retain a \textit{constant-sized} memory while enabling online processing for spatio-temporal reasoning problems. Our network maintains a memory with  a fixed number of separate memory slots which are updated through time (Fig. \ref{fig:overall framework}). Our main novelty is an algorithm based on Gumbel-Softmax that learns the memory update strategy based on the new input information and current memory slots. This strategy helps us locate the memory slot that will be discarded automatically and maintain constant memory with minimal loss of information. 
This kind of memory structure has more potential to scale up to long videos beyond a few seconds. 
In this paper, we apply it to practical tasks such as VOS and video prediction. We believe that it
can also be applicable to many spatio-temporal reasoning problems.

Our contributions can be summarized as follows:
\begin{itemize}
    \item We introduce a novel space-time memory network which has a constant number of slots. The proposed memory network can handle a long sequence of images efficiently without increasing the size of the memory when the length of the image sequence grows.
    \item We propose a novel memory update module which can be trained to select one of the memory slots for deletion and update it with the new information in the new input frame after deletion.  We show that the learned update module outperforms other hand-engineered memory update rules.
    \item We evaluate the proposed memory model on two important spatio-temporal reasoning tasks: VOS and video prediction. 
    We obtain  state-of-the-art performance with the proposed memory model, outperforming recent transformer-based models on the VOS task as well as other recent methods on the video prediction task. 
\end{itemize}

\section{Related Work}
\textbf{Implicit Memory.} Long short-term memory (LSTM) \cite{LSTM} improves over vanilla recurrent networks (RNN) by allowing the cell state (memory) to stay constant if there is no additional input. It also includes input and forget gates which allow information to be filtered before being used to update the memory. 
These gates are usually implemented with fully connected layers and hence cannot be applied to spatial-temporal sequences. \cite{shi2015convolutional} replace the linear layers with convolutional layers instead hence generating spatially-indexed memory that can be used for spatio-temporal inference. PredRNN~\cite{wang2017predrnn} modifies RNN by allowing the memory state to flow vertically through stacked RNN layers and horizontally through time. Causal LSTM~\cite{wang2018predrnn++} improves upon PredRNN by proposing the Highway Unit allowing the gradient to flow quickly to long-range input. E3D-LSTM~\cite{wang2018eidetic} uses 3d conv and attention module to improve the RNN. MIM~\cite{wang2019memory} models the stationary and non-stationary properties in the spatial-temporal dynamics by exploiting differential signal between adjacent recurrent states. However, most of these architectures only have a single memory vector/tensor which struggles to capture multiple appearances of objects in video reasoning problems ~\cite{kim2018multi}.

\textbf{Explicit Memory.} Neural Turing Machine (NTM) \cite{NTM} extended the traditional RNN with an external memory. 
NTM architecture  includes a controller and a memory bank. These two components interact with each other through multiple read and write heads similar to the attention model in transformers. The controller receives information, interacts with the memory through read and write heads to retrieve related information, remove old information, and save new ones if useful. Based on the retrieved information, a corresponding response is returned.
However, it is non-trivial to scale NTM's memory for visual tasks which require spatial information. 


\textbf{Space Time Memory (STM) and other variants.} 
STM \cite{STM} is a network based on the transformer idea that utilizes an convolutional encoder to encode the features of each frame to 2 tensors, key and value. 
In the reading stage, STM matches the query frame and template frames at potentially different spatial locations. The computed attention score are then used for a weighted sum of the features from each frame. STM does not update existing memory slots and simply concatenates the feature tensors of each new frame. One advantage of this simple mechanism is that the information on each template is not contaminated by other ones. However, it is potentially time and memory consuming since old templates are never removed and the memory grows linearly w.r.t. the number of frames. To reduce the 
time/memory costs, STM uses a simple heuristic to add a new template every 5 frames, but its time/memory consumption can still be a significant issue in longer sequences. STCN~\cite{cheng2021stcn} improves STM by proposing a more efficient feature extractor and a change to the attention mechanism (i.e. using the L2 similarity instead of cosine similarity). However, the memory problem still persists in STCN.

\textbf{Transformer with Efficient Memory for VOS.} GC~\cite{Global_context_module} compresses the memory of STM by constructing a global memory by averaging the memory of past frames and rearranging the order of the attention operation. While GC maintains a constant-size memory like our method, its performance on VOS is not as strong as the original STM's performance possibly due to the use of simple average operation. PAM \cite{Wang_2021_CVPR} also introduces a compact memory representation for the STM framework. Memory updates are performed only when significant changes between frames exist to avoid updating the memory with redundant information. Also, instead of performing memory updates with full images, pixel-wise memory updates are utilized in order to store only relevant information to the target object in the memory. Although PAM builds more efficient memory representations than STM does, its memory size still increases over time as there is no deletion operation. In contrast to PAM, our method maintains a fixed-size memory regardless of the video length, while still achieving competitive performance.

\textbf{Transformer with Efficient Memory for Other Tasks.}
Other transformer variants with efficient memory have been proposed for NLP and other vision tasks. \cite{dai-etal-2019-transformer} improves upon the transformer network by storing previous hidden states in a fixed-size queue. The queue helps the model to trade-off the context length for the memory size. \cite{Rae2020Compressive} trade-offs the granularity of the memory for the context length by compressing multiple hidden states before adding them to the memory. Therefore, the queue can keep more information, given the same size queue as in \cite{dai-etal-2019-transformer}. 
\cite{lei2020mart} augments the transformer model with a recurrent memory which utilizes a fixed number of memory slots. The proposed memory is similar to ours in the sense that it also adopts the recurrent architecture with the fixed number of memory slots. However, their target task is video captioning, so the memory is not spatially indexed.
Note that spatially-indexed memory models are significantly larger hence one cannot afford to fully-connect the memory and learn all the updates which means network design becomes more important.
\cite{Adrian2021} proposes an efficient video transformer model that utilizes local space-time attention with a fixed-size temporal window and demonstrates that the proposed model is much more efficient than a video transformer that utilizes full space-time attention, while still being effective at capturing long-term dependencies for the video classification task. Our work is different from this work in the sense that we model long-term information by maintaining and updating the fixed-size memory over time rather than modeling it through multiple local space-time attention operations.

\section{Methods}


\subsection{Overview}
Our framework includes a Query Encoder ($QE$), a Memory Encoder ($ME$), a Decoder ($D$), and a fixed-size memory $Mem = \{X^M_1, X^M_2, \dots, X^M_{K}\}$ as shown in Fig.~\ref{fig:overall framework}. $QE$ takes the query image $I_t$ as input and outputs the query feature map $X^q_t$. $QE$ can also serve as $ME$ if $ME$ does not take any additional data as input. In many VOS algorithms (but not all), $ME$ also takes an object mask $M_t$ as input in addition to the image, so a separate network should be set for $ME$. In this case, $ME$ takes $(I_t, M_t)$ as input and produces a spatially-indexed feature tensor $X^M_t$. Both $QE$ and $ME$ are implemented using convolutional neural networks (CNN). The new memory feature $X^M_t$ at each step is fed into the memory update process (Sec. \ref{sec: memory update}). 
To generate the output at step $t+1$, the query feature $X^q_{t+1}$ is matched with the memory using attention-based operation described in Sec. \ref{sec: inference stage}. The readout feature map is decoded to desired output using a Decoder implemented with upsampling CNN layers similar to U-Net\cite{Ronneberger2015UNetCN}. 
The following subsections describe our memory update and inference process in detail.

\begin{figure*}[ht]
\begin{center}
\includegraphics[width=.9\linewidth]{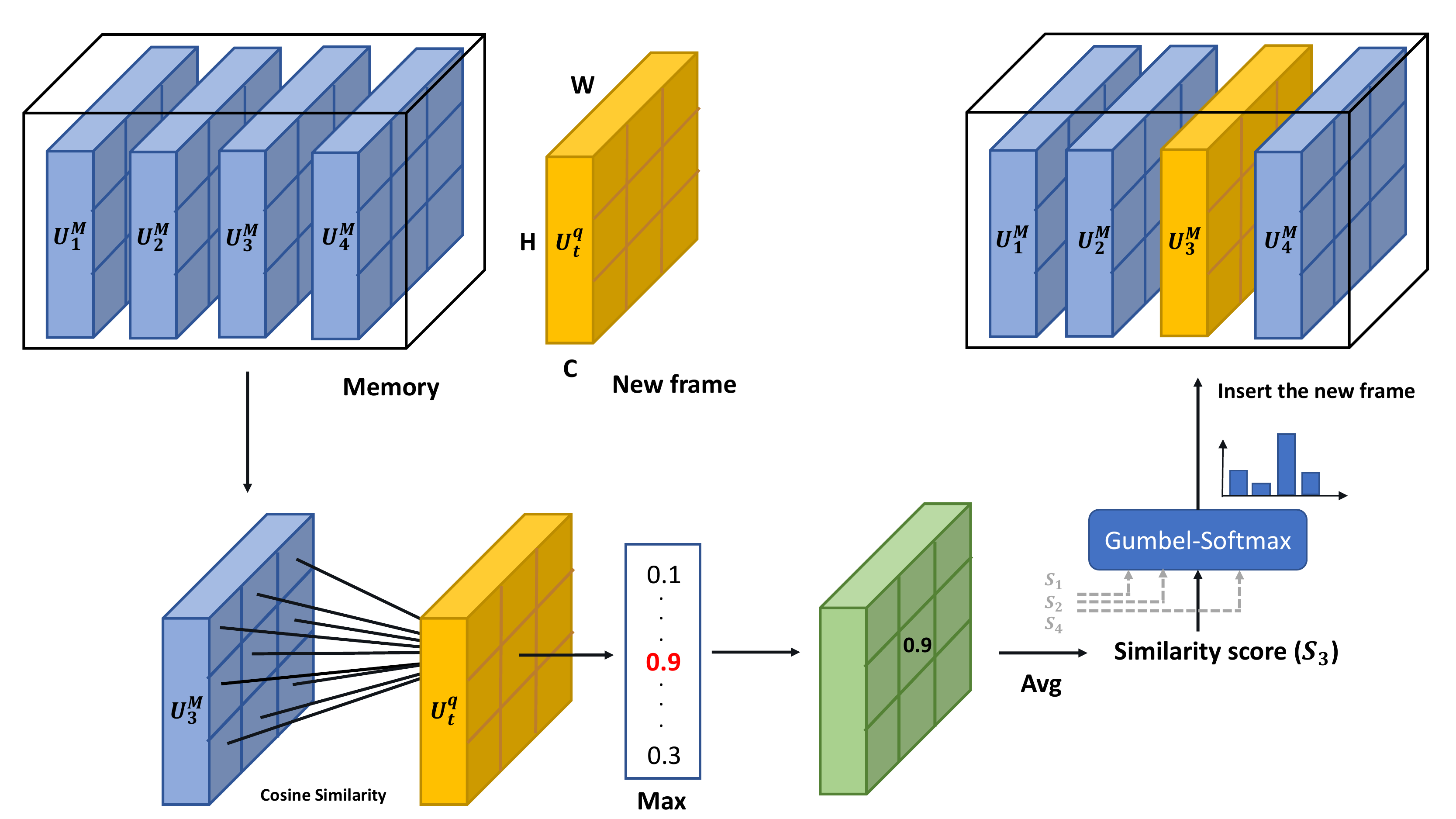}
\end{center}
\vskip -0.25in
\caption{Proposed memory update module}\label{fig:similarity_score}
\vskip -0.25in
\end{figure*}

\subsection{Memory Update Process} \label{sec: memory update}
Let $K$ and $n$ be the maximum number and the current number of slots in the memory. If $n < K$, new templates are simply added to the memory. If $n = K$, then a template must be removed from one of the memory slots to create an empty slot for the new template. One way to select a slot is to use a rule-based approach. However, there are many different rules that one can choose, so figuring out the best rule is not trivial, and the best rule may also be dependent on the task. In this work, instead of relying on hand-engineered rules, we propose to learn the selection operation via attention. The attention weights over memory slots are generated by a network that takes the new input frame and the current memory as input (Fig. \ref{fig:similarity_score}). By learning the selection operation, we remove the need for manually finding the best rules for different tasks and instead let the network to come up with the best one during training.

We obtain attention weights over memory slots as follows. Given the new template $X^q_t$ (i.e., query feature map) and stored templates $X^M_1, X^M_2, \dots, X^M_{K}$ in the memory, we learn convolutional layers to convert the query feature map $X^Q_t$ into a new \textit{update key} map $U^Q_t$ and convert the memory feature maps $X^M$ into the new update key maps $U^M$ respectively. We then compute the similarity score between  $U^Q_t$ and each of $U^M_1, U^M_2, \dots, U^M_{K}$ by using the following similarity function:
\begin{equation}
\label{eq:matching}
    s_i = S(U^Q_t, U^M_i) =  \frac{1}{HW} \sum_p \max_q  \frac{U^Q_t(p)\cdot U^M_i(q)}{||U^Q_t(p)|| ||U^M_i(q)||}
\end{equation}
where $s_i$ is the similarity score between the stored template in the $i$th memory slot and the new template, and $U_i(p)$ is the update key map of the $i$-th slot at the spatial location $p$. Note that this similarity function offers maximal flexibility in terms of matching, allowing pixels from anywhere in the images to match with each other. Such flexibility is necessary when the object is stored at different locations in different templates due to the motion and deformation of the object. Because of the encoding step, feature maps such as $U^Q$ and $U_i^M$ would have already encoded local shape and texture information, so each pixel of such feature maps is likely to represent a part of the object. Hence, intuitively, Eq. (\ref{eq:matching}) allows deformable matching over different parts of the object.

Once we compute all the similarity scores $s_1, s_2, \dots, s_{K}$ for $K$ templates in the memory, we obtain the attention vector by using Gumbel Softmax~\cite{jang2017categorical}:
\begin{equation}
\label{eq:gumbel_softmax}
    \mathbf{a}^{soft} = [a^{soft}_1, a^{soft}_2, \dots, a^{soft}_{K}] = \text{Gumbel-Softmax}([s_1, s_2, \dots, s_{K}])
\end{equation}
\begin{equation}
\label{eq:gumbel_softmax_formula}
    a^{soft}_i=\frac{exp((s_i+g_i)/\tau)}{\Sigma_{j=1}^K exp((s_j+g_j)/\tau)}
\end{equation}
where $g_1...g_k$ are iid sampled from the Gumbel(0,1) distribution, and $\tau$ is a hyperparameter which is set to $1$ in our experiments.
We utilize Straight-Through Gumbel-Softmax~\cite{jang2017categorical} which converts $\mathbf{a}^{soft}$ to a one-hot vector $\mathbf{a}^{hard}$ with the argmax function in the forward pass and still allows gradients to be computed by using $\mathbf{a}^{soft}$ in the backward pass. 
Note that the update key feature maps are learned, 
which allows the network to avoid always selecting a slot that stores the most similar appearance to that of the input feature map $X^q_t$ for deletion. Once we obtain the hard attention vector $\mathbf{a}^{hard}$, we update each memory slot as follows:
\begin{equation}
\label{eq:memory_update}
    X^M_i = (1-a^{hard}_i)X^M_i + a^{hard}_iX^q_t.
\end{equation}
Since $\mathbf{a}^{hard}$ is a one-hot vector in the forward pass, Eq. (\ref{eq:memory_update}) allows us to remove the template in the selected slot completely and add the new template to the slot.

Note that our update process is different from the memory update processes in other memory networks such as the Gated Recurrent Units (GRU) ~\cite{GRU} or Neural Turing Machines (NTM) ~\cite{NTM} in two ways. First, in both GRU and NTM, soft attention weights are used to mix the information of the current input with the information of the memory during the memory update. Second, these approaches selectively take the information in the current input by using the attention weights defined over the input. In contrast, we dedicate a new memory slot for the input $X_t$ as shown in Eq. (\ref{eq:memory_update}). An intuition for this choice is that the new information is fresh and might be the most relevant to instant predictions. Also, in our framework, it is much easier to understand what kind of information that the model keeps in the memory because our memory stores the selected frames in the memory rather than fusing all the information across the frames into the memory.

As for the information fusion across the frames, we have actually implemented a fusion module in which the template selected for deletion is fused to other templates in the memory before its deletion. 
However, in our experiments, the model with the fusion module did not outperform the model without the fusion module. Please refer to the supplementary document for more details about the fusion module we implemented and the results we obtained.

\subsection{Inference Stage} \label{sec: inference stage}
 
In the inference stage, the query frames go through the Query Encoder to be encoded into a feature map and converted into a key-value pair ($k^Q$, $v^Q$). 
Similarly, each template in the memory is converted into a key-value pair, and all templates are  concatenated to form ($k^M$, $v^M$). $k^Q$ and $v^Q$ contain information from the query frame, and $k^M$ and $v^M$ carry information of the past frames that are currently stored in the memory. Then each pixel $i$ of the query template is densely matched with every pixel $j$ of memory's templates through the attention mechanism similar to \cite{STM}:
\begin{equation}
    y_i = \left[\frac{1}{\Sigma_j \exp(k_i^{Q\top}  k^M_j)} \exp(k_i^{Q\top} k^M_j)v^M_j, v^Q_i\right]
\end{equation}
where $1\leq i \leq HW$, $1\leq j \leq KHW$, and $[\cdot, \cdot]$ is the concatenation. \textbf{y} is then decoded into the output with a standard U-Net decoder without highway connections. Note that the decoder receive information from the encoder only through the memory, which reduced storage costs during both the training and inference times.

\section{Experiments} \label{experiment}
We conduct experiments on two difficult video tasks, video object segmentation and video prediction. They are described in the subsequent subsections.
\subsection{Video object segmentation} \label{sec: VOS}



The VOS problem is chosen as a benchmark because of following reasons. First, VOS requires both long and short term memory to segment the objects efficiently. Second, the variation of appearance of the object throughout the video is significant, which requires the network to keep a diverse memory. Finally, the segmentation task requires the network to preserve a huge number of spatial details, which is a challenging task for memory networks and a spatially-indexed memory is necessary. 

\subsubsection{Implementation Details}

The architectures of the encoder and decoder are the same as those of \cite{STM}. However, we replace batch normalization~\cite{ioffe2015batch} with group normalization~\cite{wu2018group} in our backbone - Resnet 50~\cite{he2016deep}. This makes the training more stable because of the small batch size that we use due to memory constraints. 
The network uses 6 memory blocks including: 1 block for the ground truth (1st frame), 1 block for the latest frame, and 4 blocks for intermediate frames.
Post-processing is also employed to remove some false positive predictions that are far away from the object in the previous frame. Please refer to the supplementary document to see the details of our post-processing. 

\textbf{Training.} The model is trained in two stages: pre-training with images and fine-tuning with videos.  In the pre-training stage, consistent with prior work~\cite{STM}, we create short clips consisting 5 frames by using images from COCO~\cite{lin2014microsoft}, ECSSD~\cite{ecssd}, SBD~\cite{sbd}, MSRA10k~\cite{msra10k}, and HKUIS~\cite{hkuis}. We minimize the cross entropy loss using the Adam optimizer~\cite{KingmaB14} with learning rate $10^{-4}$. In the fine-tuning stage, we randomly extract video clips consisting of 10 frames from videos in the DAVIS and Youtube training set. We fine-tune the model with the Adam optimizer with the cosine learning rate scheme between $[10^{-5}, 10^{-7}]$. We use $384\times384$ as the input image resolution in both stages.
    

\subsubsection{Ablation Experiments.} \label{sec: memory analysis}
We first compare our learned memory update module with rule-based memory update modules. We implement the following memory update rules as our baselines.

\begin{itemize}
\vspace{-0.02in}
    \item \textbf{A}: 
    Remove the oldest template in the memory (queue). 
\vspace{-0.02in}
    \item \textbf{B}: 
    Remove the newest template in the memory (stack). 
\vspace{-0.02in}
    \item \textbf{C}: Drop one of the templates in the memory randomly.
\vspace{-0.02in}
    \item \textbf{D}: Select $K$ frames randomly among all the previous frames and store the corresponding templates in the memory.
    \item \textbf{E}: Keep only the first and the most recent frame in the memory.
    \item \textbf{F}: Remove the template that stores the most similar appearance to that of the newest template.
\vspace{-0.02in}
\end{itemize}

Table \ref{tab: memory variant} shows the results of our ablation experiments. The proposed model outperforms other baselines that are implemented with the rule-based memory update modules. This shows that the network can learn what to keep in the memory effectively when the number of memory slots is limited.

\begin{table}[htb]
\caption{Results on DAVIS test-dev with different memory update strategies.}
\label{tab: memory variant}
\centering

\begin{tabular}{l|c|c|c}
\textbf{Memory models}      & \textbf{J }   & \textbf{F}    & \textbf{Mean}    \\ 
\hline

(\textbf{A}) Remove oldest    &73.3 & 80.0 & 76.7\\
(\textbf{B}) Remove newest   &72.9 &79.2 & 76.0\\
(\textbf{C}) Random drop  & 72.4 & 78.7 & 75.5  \\
(\textbf{D}) Random select  & 73.7 & 80.1 & 76.9 \\
(\textbf{E}) Keep the first and last frame  & 71.5 & 78.8 & 75.2 \\
(\textbf{F}) Drop the most similar template  & 73.5 & 80.2 & 76.9 \\
(\textbf{Ours}) Learned & \textbf{74.8} & \textbf{81.8} & \textbf{78.3} \\
\hline
\end{tabular}

\end{table}

We also report the model performance with different numbers of memory slots. As shown in Table \ref{tab: num_slot}, increasing the number of the slots leads to better performance in general. For the VOS task, we achieve the best performance with 6 slots.  

\begin{table}
\caption{Results on DAVIS test-dev with different number of memory slots.}
\label{tab: num_slot}
\centering
\begin{tabular}{c|c|c|c|c|c}
\textbf{Num slots} & \textbf{3} & \textbf{4} & \textbf{5} & \textbf{6}    & \textbf{7} \\ \hline
\textbf{J\&F}      & 76.3       & 76.9       & 77.8       & \textbf{78.3} & 77.3       \\ \hline
\end{tabular}
\end{table}



\subsubsection{Results on the DAVIS 2017 and Youtube 2018 Dataset.}

\begin{table} [htb]

\centering
\caption{Results on the DAVIS 2017  dataset while fine-tuning on both DAVIS and  Youtube.}
\label{tab: davis_youtube_training}
\vskip -0.1in
\scalebox{0.9}{
\begin{tabular}{l|c|c|c|c|c|c|c|c}
\textbf{}       & \textbf{}         & \multicolumn{1}{l|}{\textbf{}} & \multicolumn{3}{c|}{\textbf{Validation}} & \multicolumn{3}{c}{\textbf{Test-dev}} \\ \cline{4-9} 
\textbf{Method} & \textbf{Backbone} & \textbf{Memory size}           & \textbf{J}  & \textbf{F}  & \textbf{Avg} & \textbf{J} & \textbf{F} & \textbf{Avg} \\ \hline
STM~\cite{STM}        & Resnet 50         & linear growth                  & 79.2        & 84.3        & 81.8         & 69.3       & 75.2       & 72.2         \\
STM~\cite{STM}        & Resnet 50         &    constant               & 77.7        & -        & -         & -       & -       & -         \\
KMN~\cite{seong2020kernelized}       & Resnet 50         & linear growth                  & 80.0        & 85.6        & 82.8         & 74.1       & 80.3       & 77.2         \\
EGM~\cite{episodic_graph_memory}       & Resnet 50         & linear growth                  & 80.2        & 85.2        & 82.8         & -          & -          & -            \\
CFBI~\cite{yang2020CFBI}       & Resnet 101        & N/A                            & 79.1        & 84.6        & 81.9         & 71.1       & 78.5       & 74.8         \\
SwiftNet~\cite{Wang_2021_CVPR}       & Resnet 50        & linear                            & 78.3        & 83.9        & 81.1         & -       & -       & -         \\
A-Game~\cite{AGame_2019_CVPR}     & Resnet 101        & N/A                            & 67.2        & 72.7        & 70           & -          & -          & -            \\
FEELVOS~\cite{Voigtlaender2019FEELVOSFE}    & DeepLab v3        & N/A                            & 69.1        & 74.0        & 71.5         & 55.1       & 60.4       & 57.8         \\ 
PAM~\cite{Wang_2021_CVPR} & Resnet 50 & linear growth & 78.3 & 83.9 & 81.1 & - & - & -\\
RMNet~\cite{xie2021efficient} & Resnet 50 & linear growth &81.0 & 86.0 & 83.5 &71.9 &78.1&75.0\\ 
LCM~\cite{Hu_2021_CVPR} & Resnet 50 & linear growth &80.5 & 86.5 & 83.5 &74.4 & \textcolor{blue}{81.8} & 78.1\\
STCN~\cite{cheng2021stcn} & Resnet 50 & linear growth & \textcolor{blue}{82.2} & \textbf{88.6} & \textcolor{blue}{85.4} &73.1 & 80.0 & 76.5\\
\hline

STReMN & Resnet 50 & constant & 81.3 & 86.1 & 83.7 & \textcolor{blue}{74.8} & \textcolor{blue}{81.8} & \textcolor{blue}{78.3}\\
STReMN-MS & Resnet 50 & constant & \textbf{83.4} & \textcolor{blue}{88.2} & \textbf{85.8} & \textbf{75.6} &\textbf{ 82.7} & \textbf{79.1}\\
\hline

\end{tabular}
}
\end{table}

\vskip -0.1in

\begin{table}
\centering
\caption{Results on Youtube VOS 2018 validation set.  \textbf{R50/101}: use resnet 50/101 as backbone.}
\label{tab: Youtube18}
\scalebox{0.90}{
\begin{tabular}{l|cc|cc|c}
\textbf{Methods} & \multicolumn{2}{c|}{\textbf{Seen}} & \multicolumn{2}{c|}{\textbf{Unseen}} &   \\ \hline
        & \textbf{J}           & \textbf{F}           & \textbf{J}            & \textbf{F}            &  \textbf{Avg}    \\ \hline

GC (R50)~\cite{Global_context_module}      & 72.6        & 75.6        & 68.9         & 75.7         & 73.2 \\
EGM(R50)~\cite{episodic_graph_memory}     & 80.7        & 85.1        & 74.0         & 80.9         & 80.2 \\
CFBI(R101)~\cite{yang2020CFBI}    & 81.1        & 85.8        & 75.3         & 83.4         & 81.4 \\
PReMVOS(R101)~\cite{luiten2018premvos} & 71.4        & 75.9        & 56.5         & 63.7         & 66.9 \\
A-Game(R101)~\cite{AGame_2019_CVPR}  & 67.8        & -           & 60.8         & -            & 66.1 \\ 
KMN (R50)~\cite{seong2020kernelized} & 81.4 & 85.6 & 72.8 & 84.2 & 80.9 \\ 
STM(R50)~\cite{STM}     & 79.7        & 84.2        & 72.8         & 80.9         & 79.4 \\ 
LCM(R50)~\cite{Hu_2021_CVPR}     & \textbf{82.2}        & \textbf{86.7}        & 75.7         & 83.4        & 82.0\\
PAM(R50)~\cite{Wang_2021_CVPR}     &77.8        & 81.8        & 72.3         & 79.5         & 77.8 \\
RMNet(R50)~\cite{xie2021efficient}     &\textcolor{blue}{82.1}        & 85.7        & 75.7         & 82.4         & 81.5 \\
STCN(R50)~\cite{cheng2021stcn}     &81.9        & \textcolor{blue}{86.5}        & \textbf{77.9}         & \textbf{85.7}         & \textbf{83.0} \\
\hline

STReMN (R50)     & 79.6       & 83.8       & 76.1        & 84.1       & 80.9\\
STReMN-MS (R50)    & 80.7       & 85.0       & \textcolor{blue}{77.6}        & \textcolor{blue}{85.5}       & \textcolor{blue}{82.2}\\
\hline

\end{tabular}
}
\end{table}

DAVIS 2017 includes 60 videos, 30 videos, and 30 videos in the training, validation, and test-dev sets respectively. 
The objects range from common classes such as human or dog to rare classes such as string, car wheel, and drone. The number of frames in each video can be up to 130 frames. Youtube 2018 includes 3471 training videos with dense object annotations at 6 fps and 474 validation videos which comprise of both seen and unseen object categories in the training set. The length of each video can be up to 180 frames. The methods are evaluated using the mean region similarity J and the mean contour accuracy F~\cite{davis2017}. 

Note that we only compare our method with other methods that are trained on the training set and tested on the validation and testing sets. The  online fine-tuning methods~\cite{perazzi2017learning}\cite{caelles2017one}\cite{voigtlaender2017online}\cite{bhat2020learning}\cite{bao2018cnn}\cite{li2018video}\cite{hu2018maskrnn}\cite{robinson2020learning}, which fine-tune the models on each testing video separately, are  not general memory models and hence orthogonal to our work. 
We also exclude the AOT approach \cite{yang2021aot} in our comparison as their contributions are in utilizing local attention which allows the model to better focus on the foreground regions and combining multiple single object predictions efficiently, which is orthogonal to our contributions of designing constant-size memory. Since our method is general, it can be plugged into other methods such as \cite{yang2021aot} in order to address the memory size issue for long videos.

We present two versions of STReMN. We implement STReMN with the STM backbone and our proposed memory mechanism. We also use the Atrous Spatial Pyramid Pooling (ASPP) layer \cite{chen2018encoder} on top of the memory-decoded feature map. We implement STReMN-MS with the multiscale testing from CFBI~\cite{yang2020CFBI}. Table~\ref{tab: davis_youtube_training} shows our results on the DAVIS 2017 validation and test-dev sets, when fine-tuned on both the DAVIS and the YouTube datasets. As shown in Table~\ref{tab: davis_youtube_training}, STReMN matches the performance of all the linear-size memory models on the validation set and is better than several other networks with larger backbone and memory. STReMN-MS achieves SOTA performance on both the DAVIS validation and test-dev set.

Table~\ref{tab: Youtube18} shows our result on the Youtube 2018 validation set. With marginal performance difference to the best performing one, STCN, our best model ranks second both on average and on objects unseen during the training. We believe that our performance on this dataset is still very competitive compared to that of STCN, considering our method utilizes constant-size memory rather than linear-size memory used in STCN.

\subsubsection{Qualitative Results.}

\begin{figure}
\begin{center}
\includegraphics[width=0.7\linewidth]{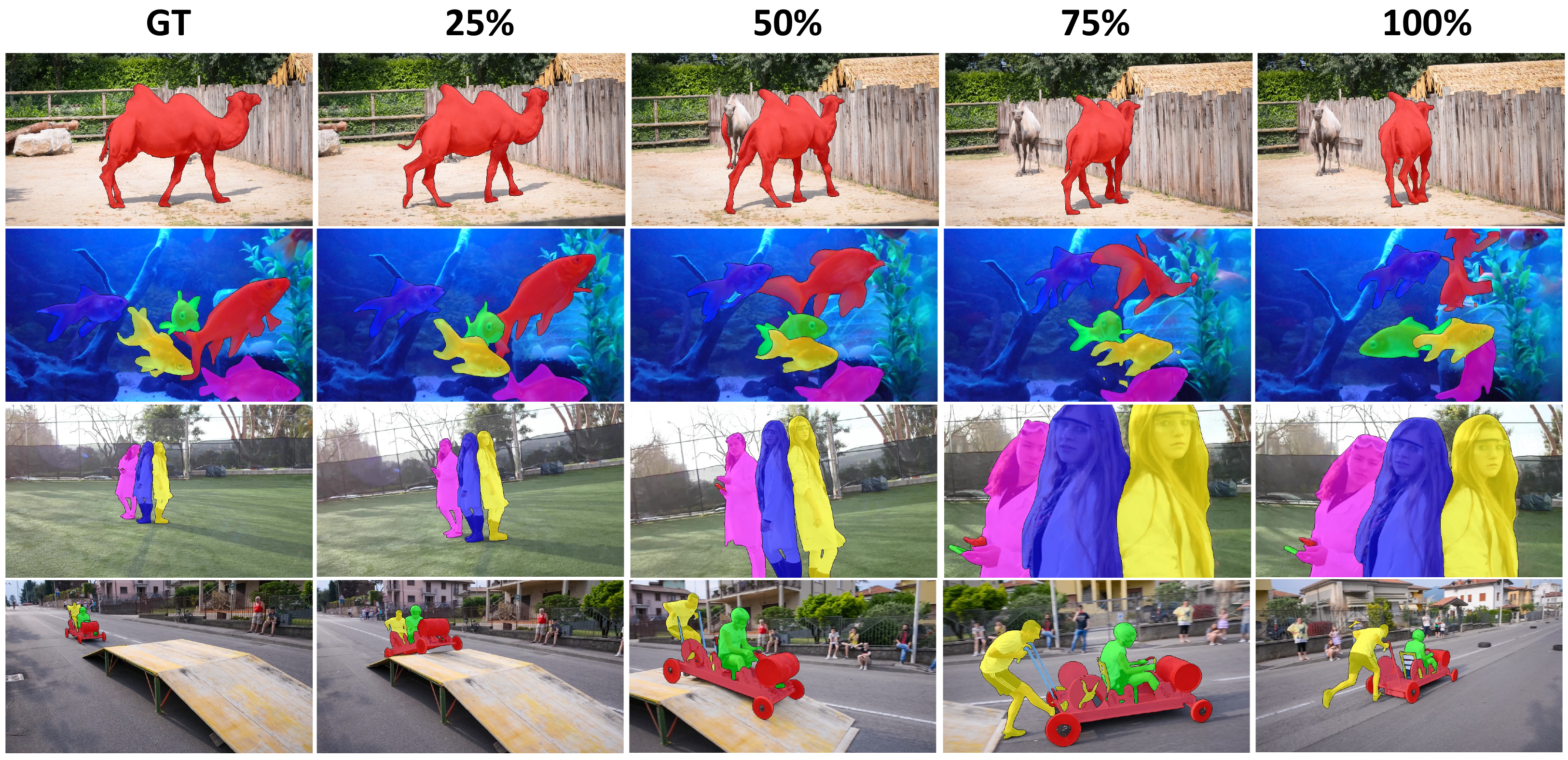}
\end{center}
\vskip -0.3in
   \caption{Qualitative results (Best viewed in color).}
\vskip -0.15in
\label{fig: good_viz}
\end{figure}

In Fig.~\ref{fig: good_viz}, 
we show some qualitative results on the DAVIS 2017 dataset. The results show that our model is able to capture a variety of object poses, maintain long-term memory after partial occlusion, and adapt well to most of the appearance changes. However, our model still fails to retrieve extremely small objects such as the phones held by the pink girl in some frames. Besides, on the camel video, the model can ignore most but not completely the second camel. This suggests that the model could not completely grasp the concept of a single object. Other potential solutions are adding more global context by using self attention or designing a better decoder to remove these background noises.  More visual examples are discussed in the supplementary material.

\subsection{Video prediction}\label{sec: video prediction}
Another task we chose to benchmark our approach is video prediction. In this task, the model needs to predict $N$ frames in the future given the first $M$ observed frames. Therefore, this task requires the memory to preserve both the spatial information (e.g. object appearance and background texture) and the temporal information (e.g. motion pattern).

\noindent \textbf{Implementation details.} As no ground truth segmentation mask is available, we  use the same encoder for both the query and the memory. The input at step $t$ is the stack of 3 consecutive video frames $(I_{t-3}, I_{t-2}, I_{t-1})$. The stacked video frames allows the memory to keep both appearance and motion patterns. The encoder is implemented with 6 CNN blocks. Each block consists of 2 CNN layers followed by group normalization and Leaky RELU. The decoder includes 6 deconvolution  blocks. Each block has 2 deconvolution layers followed by group normalization and leaky RELU. The memory module is the same as in the model for VOS and includes 5 slots.

\noindent \textbf{Training.} For a fair comparison, our model is not pretrained on other datasets. The model is trained using Adam optimizer with learning rate set to $10^{-3}$ to minimize the shrinkage loss~\cite{lu2018deep} on $L_1$ and $L_2$ and the scale invariant gradient loss~\cite{ummenhofer2017demon}. The final model is obtained by averaging model weight on different timesteps having high SSIM scores.

\noindent \textbf{Human3.6M}~\cite{h36m_pami}: The Human3.6M dataset captures complex motion from general human actions. We follow~\cite{Guen_2020_CVPR} to use the walking videos with scene S1, S5, S6, S7, and S8 for training and S9 and S11 for testing. Each video is resized to $128\times 128$. In both training and testing, the model predicts 4 frames in the future given the first 4 observed frames. Our approach outperforms other memory models by a significant margin on 3 different metrics mean squared error (MSE), mean average error (MAE), and SSIM as shown in Table \ref{tab: video_pred}. 
Structural similarity index measure (SSIM) is a perceptual-based metric. Higher SSIM shows that our model can produce frames with better structural contents.

\noindent \textbf{KTH}~\cite{KTH}: This dataset has 25 people performing 6 types of action on 4 different scenes. The first 16 people are for training and the rest for testing. The videos are resized to $128\times 128$. In training, the model is provided with the first 10 frames of the video and trained to predict the next 20 frames. 
We achieve comparable results with PredRNN on PSNR based on MSE and SSIM (i.e. structure similarity) while outperforming it on LPIPS which measures the perceptual similarity based on the features extracted from pre-trained AlexNet. Therefore, LPIPS measures the human perception on the generated frames better than the previous 2 metrics as shown in~\cite{zhang2018unreasonable}.

We show that our model can predict perceptually reasonable feature frames in Fig.~\ref{fig: video_pred}. More results will be shown in the supplementary.

\begin{table*}
\vskip -0.2in
\begin{center}
\caption{Video prediction results.}
\label{tab: video_pred}
\scalebox{0.95}{
\begin{tabular}{l|ccc|ccc}
\textbf{}   & \multicolumn{3}{c|}{\textbf{Human 3.6}} & \multicolumn{3}{c}{\textbf{KTH}} \\ \hline 
\textbf{Method}                       & MSE/10      & MAE/100      & SSIM$\uparrow$       & \multicolumn{1}{l}{PSNR$\uparrow$} & \multicolumn{1}{l}{SSIM$\uparrow$}  & \multicolumn{1}{l}{LPIPS$\downarrow$} \\ \hline
ConvLSTM~\cite{shi2015convolutional} & 50.4 & 18.9  & 0.776 & 23.58  & 0.712& 0.231 \\
Causal LSTM~\cite{wang2018predrnn++} & 45.8 & 17.2  & 0.851      & -& - & - \\
MIM~\cite{wang2019memory}& 42.9& 17.8 & 0.79& - & -  & - \\
E3D-LSTM~\cite{wang2018eidetic}      & 46.4 & 16.6 & 0.869 & -  & -& - \\
PredRNN~\cite{wang2017predrnn}       & 48.4& 18.9 & 0.781  & 27.55   & \textbf{0.839 } & 0.204 \\
MCnet + Residual~\cite{villegas17mcnet}                                       & - & - & - & 26.29 & 0.806 & -   \\
TrajGRU~\cite{NIPS2017_a6db4ed0}& -  & -  & -  & 26.97 & 0.790  & - \\
DFN~\cite{NIPS2016_8bf1211f}            & -    & -    & -    & 27.26   & 0.794  & - \\
Conv-TT-LSTM~\cite{NEURIPS2020_9e1a3651}                                           & -           & -            & -          & \textbf{27.62}                    & 0.815                    & 0.204 \\ \hline 
\textbf{STReMN}                         & \textbf{37.89}       & \textbf{14.04}        & \textbf{0.906}      & 27.44 & 0.834 & \textbf{0.134} \\ \hline

\end{tabular}
}
\end{center}
\vskip -0.05in
\end{table*}

\begin{figure*}[ht]
\vskip -0.4in
\begin{center}
\includegraphics[width=.7\linewidth]{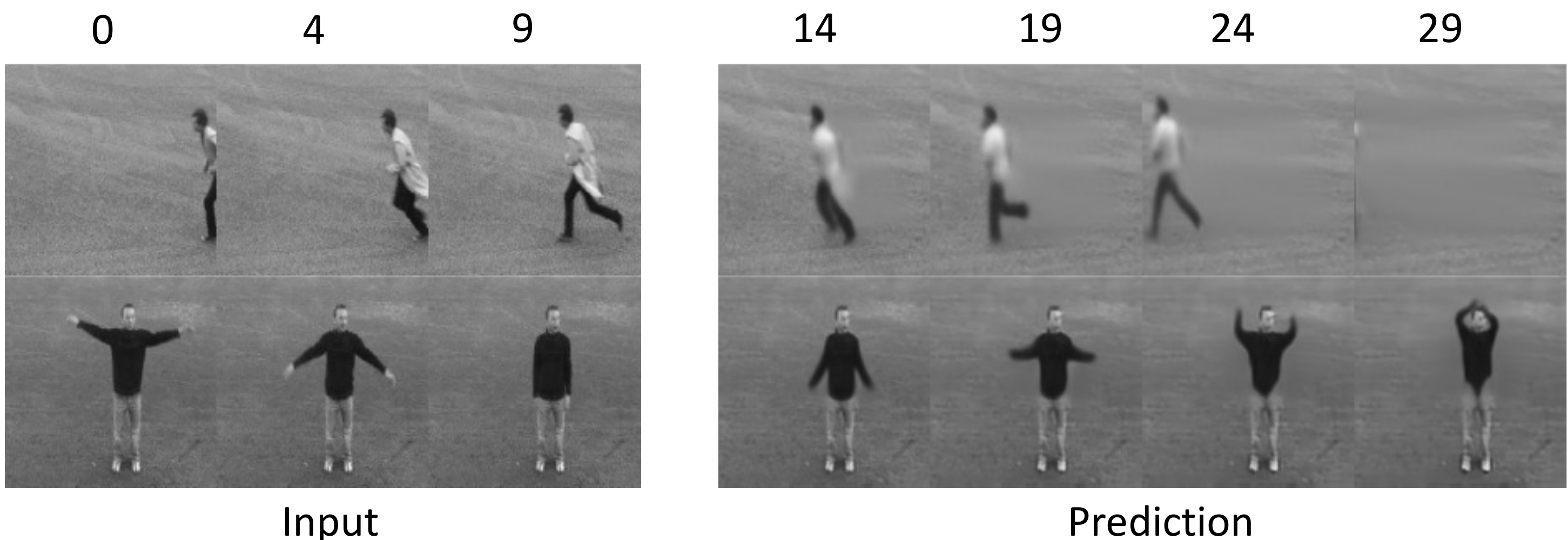}
\end{center}
\vskip -0.2in
\caption{Qualitative results for video prediction.}
\label{fig: video_pred}
\vskip -0.2in
\end{figure*}

\subsection{Memory Analysis}
\begin{figure*}[ht]
\vskip -0.3in
\begin{center}
\includegraphics[width=.7\linewidth]{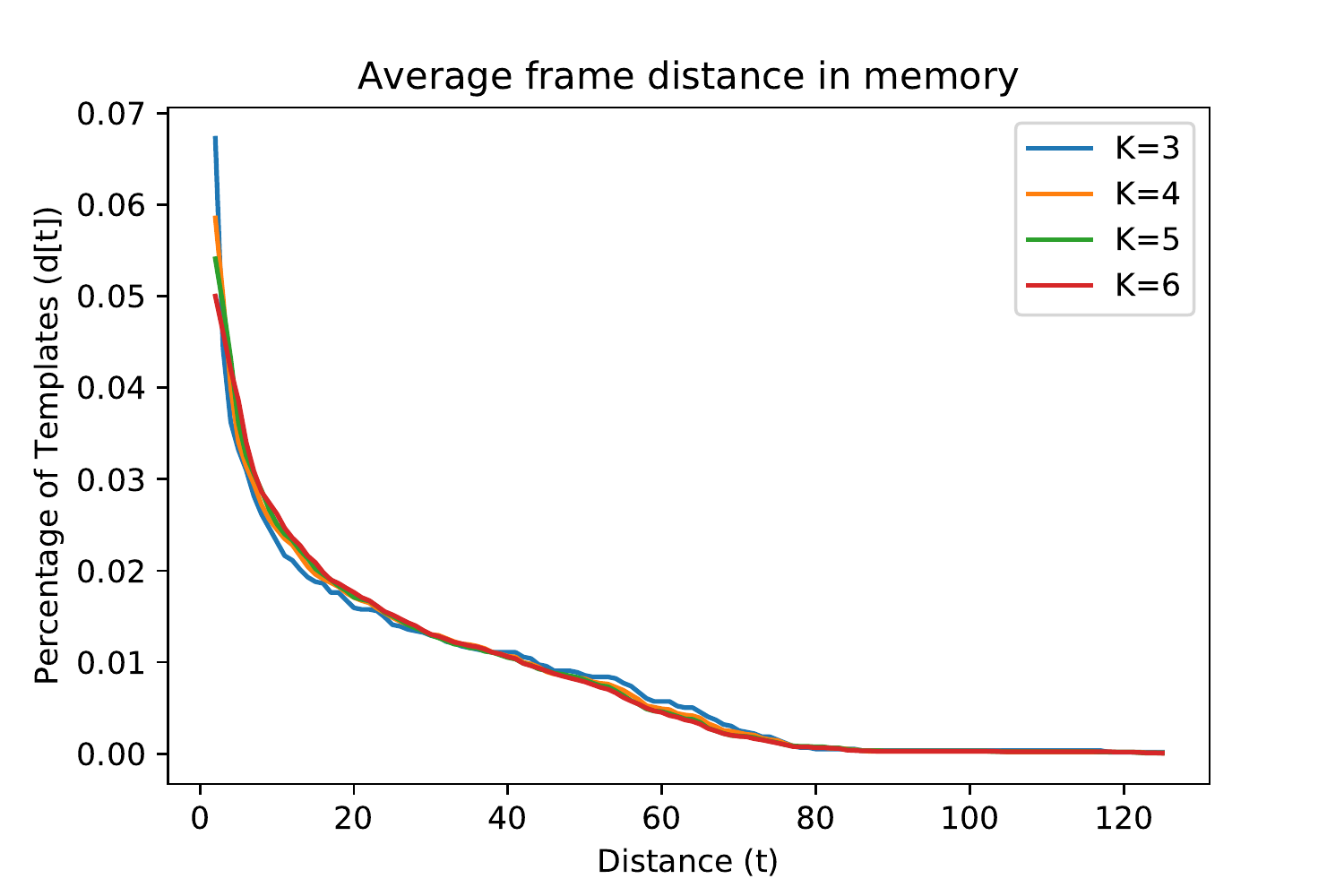}
\end{center}
\vskip -0.25in
\caption{Percentage of memory templates that appear t frames before the query.}
\label{fig: mem_distance}
\vskip -0.35in
\end{figure*}

In this section, we further investigate the behavior of the learned memory update modules in the VOS task. First, let us define $d^K[t]$ as the number of memory templates that are $t$ frames before the query frame in the memory with $K$ templates. The first frame (where the ground truth mask is provided) and the previous frame are ignored because they are always included in the memory. $d[t]$ is normalized with $\Sigma_t d^K [t]$ so that it sums to $1$. As shown in Fig.~\ref{fig: mem_distance}, the behaviors of the memory with different $K$ are consistent: they preserve the recent frames (i.e. short distance) with a higher likelihood, but maintains a sufficient amount of long-term frames (i.e. long distance). 

\begin{figure*}[ht]
\vskip -0.15in
\begin{center}
\includegraphics[width=.8\linewidth]{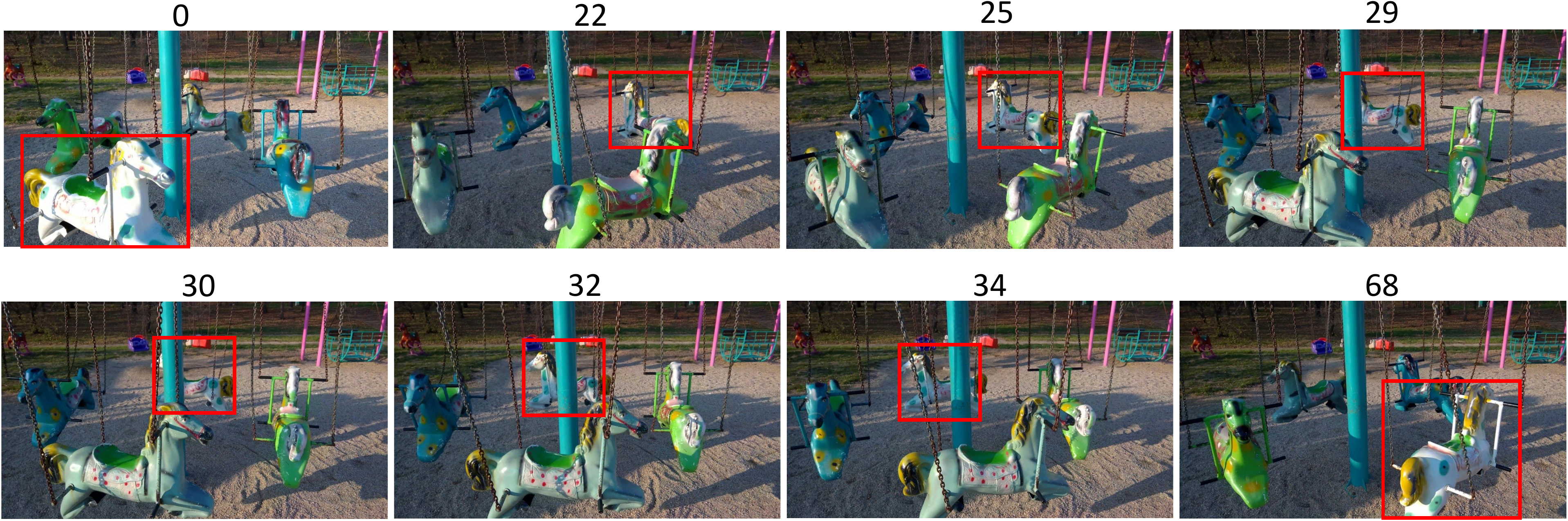}
\end{center}
 \vskip -0.15in
\caption{Carousel frames around memory templates}
\label{fig: analysis}
\vskip -0.3in
\end{figure*}

In addition, we take a look into the memory at the end of the video. Frames that were retained in the memory usually show significantly distinct appearances. For example, in the carousel video in the DAVIS test-dev dataset, at frame $68$, the memory preserves frames 0 (ground truth), 17, 18, 29, 32, and 67 (the newest frame), shown in Fig.~\ref{fig: analysis}. It can be noted that from frame 19 to 29, a shadow is cast on the horse. From frame 29 to 34, the horse is partially occluded. This shows that our memory can capture some important moments when there are significant appearance changes. 

\section{Conclusion}
In this paper, we propose a novel constant-size memory network architecture with the potential for scaling up to long videos.  
The main novelty is to learn an adaptive memory update rule using the Gumbel softmax operator. 
This allows our model to reason about information of past frames with a fixed size memory of several slots. We showed that our model worked well on two difficult tasks including video object segmentation and video prediction. We achieved state-of-the-art results on DAVIS 2017 validation and test-dev set and competitive results on  YouTube-VOS. Our results on the video prediction also showed that the proposed memory architecture outperforms other memory  architectures. In the future, we hope that our model can be extended to other tasks such as video reasoning, reinforcement learning, and embodied AI.

%
%
\bibliographystyle{splncs04}
\bibliography{egbib}

\clearpage
\appendix
\section{Running Time Analysis}
The time complexity of the space-time read for each query in STM is dominated by $O(\frac{T}{\gamma}N^2 (D_k + D_v))$ where $T$ is the number of frames before the query frame, $\gamma$ is a constant deciding the sampling frequency (i.e the memory will keep once every $\gamma$ frames in the video) for the memory ($\gamma=5$ in the STM model), $N$ is the number of pixels, and $D_k$ and $D_v$ are the dimensionality of the keys and values respectively. For STReMN, the number of memory slots is set to a \textbf{small} constant $M$ (up to $6$ in our experiments). Thus, our attention complexity is $O(MN^2 (D_k + D_v))$. Besides, to update the memory, our model first transforms the memory template and the new input. The complexity of this process is $O((M+1)ND_k^2)$, which is similar to the key-val transformation in the STM. Then the model measures the similarity between the new input and every memory template - $O(MN^2D_k)$. Therefore, the time complexity of STReMN is $(O(MN^2 (D_k + D_v) + (M+1) ND_k^2)$. Usually, the number of pixels is at a higher order than $D_k$, hence in terms of big-O we achieve linear speed-up w.r.t. STM.

In our model, the complexity for memory reading at each time step is independent on the position of the query. In STM, the complexity depends on the number of frames before the query. Therefore, STReMN is significantly faster than STM in long videos more than $50$ frames.
\section{Memory Fusion}
In this section, we investigate the effect of the fusion module that we briefly mentioned in Sec. 3.2 of the main paper. It is reasonably to think about some kind of memory fusion to combine the information from multiple slots. However in practice this is difficult for spatially-indexed memory such as ours, because one would have to first align different memory slots spatially that were potentially taken from different points of time and objects might have undergone nonlinear motion already between these times. Hence, some special designs are needed to align the features to make this fusion happen. 

Note that based on the experiment results, we \textbf{did not} utilize this module in the system, but it is presented here as a \textbf{negative result}, by which we hope to generate some food for thought for future practitioners especially because memory fusion sounds like a plausible idea. 

This module was proposed to combine the information from the deleted template with other templates in the memory. Let $X_i^M$ be the deleted template, and $X_j^M$ be the target memory template. Before being deleted, $X_i^M$ is used to update $X_j^M$. First, we apply an input filter on $X_i^M$ to get $X_i^{'M}$ as follows:

\begin{align}
    X^A_j &= FAM(X^M_i, X^M_j) \\
    FAM(X^M_i, X^M_j) &=  Softmax(X^M_i (X^M_j)^T) X^M_j \\
    r &= \sigma(f([X_i^M, X^A_j])) \\
    X'^M_i &= r \odot X^M_i 
\end{align}
where $f$ is implemented with a CNN, $\sigma$ is the sigmoid function, and $\odot$ is the Hadamard product. $FAM$, feature alignment module, is used to align pixels from $X_j^M$ to $X_i^M$ in case the target object's position changes. Finally, $X_j^M$ is updated with $X_i^{'M}$ as follows:  

\begin{align}
    X^A_i &= FAM(X^M_j, X'^M_i) \\ 
    z &= \sigma(h([X^A_i, X^M_j])) \\ 
    \hat{X}^M_j& = (1-z)\odot X^M_j + z\odot tanh(g([X^A_i, X^M_j]))
\end{align}

This was done before our proposed learning-based template updating algorithm, hence at that time we tested different strategies to choose the target template to update such as updating the least or most similar template or updating every other templates in the memory. We show the results of updating the least similar template in Table ~\ref{tab: fusion}.
The result of the "without Fusion" model was obtained by simply removed the fusion module from the model. The weight was reused without retraining. The result shows that the fusion module was trying to learn an identity function and hence unnecessary. When we used other strategies such as updating the most similar template, we obtained similar results to the one shown in Table~\ref{tab: fusion}. Hence, we finally decided to drop the fusion module based on those experiment results. 
\begin{table}[ht]
\caption{Results of the fusion module on DAVIS 17 test-dev. Note that based on this negative result, we \textbf{did not} utilize this module in the final system}
\label{tab: fusion}
\begin{center}
\begin{tabular}{c|c|c}
              & \textbf{with Fusion} & \textbf{without Fusion} \\ \hline
\textbf{J\&F} & 76.7                 & 76.7                   
\end{tabular}
\end{center}

\end{table}

It is still unclear to us whether this task would not require memory fusion because the length of the video is still not long enough (usual videos of $30-100$ frames would be well-covered with the current approach of storing a few templates and matching them) in the current benchmarks, or is it because that it is difficult to learn a good fusion module because of the alignment difficulties under deformable motion. In the future we plan to try the algorithm on very long (1000+ frames) videos and examine whether some memory fusion would bring some benefits over there.

\section{Memory Analysis}
\begin{figure*}[ht]
\begin{center}
\includegraphics[width=1.0\linewidth]{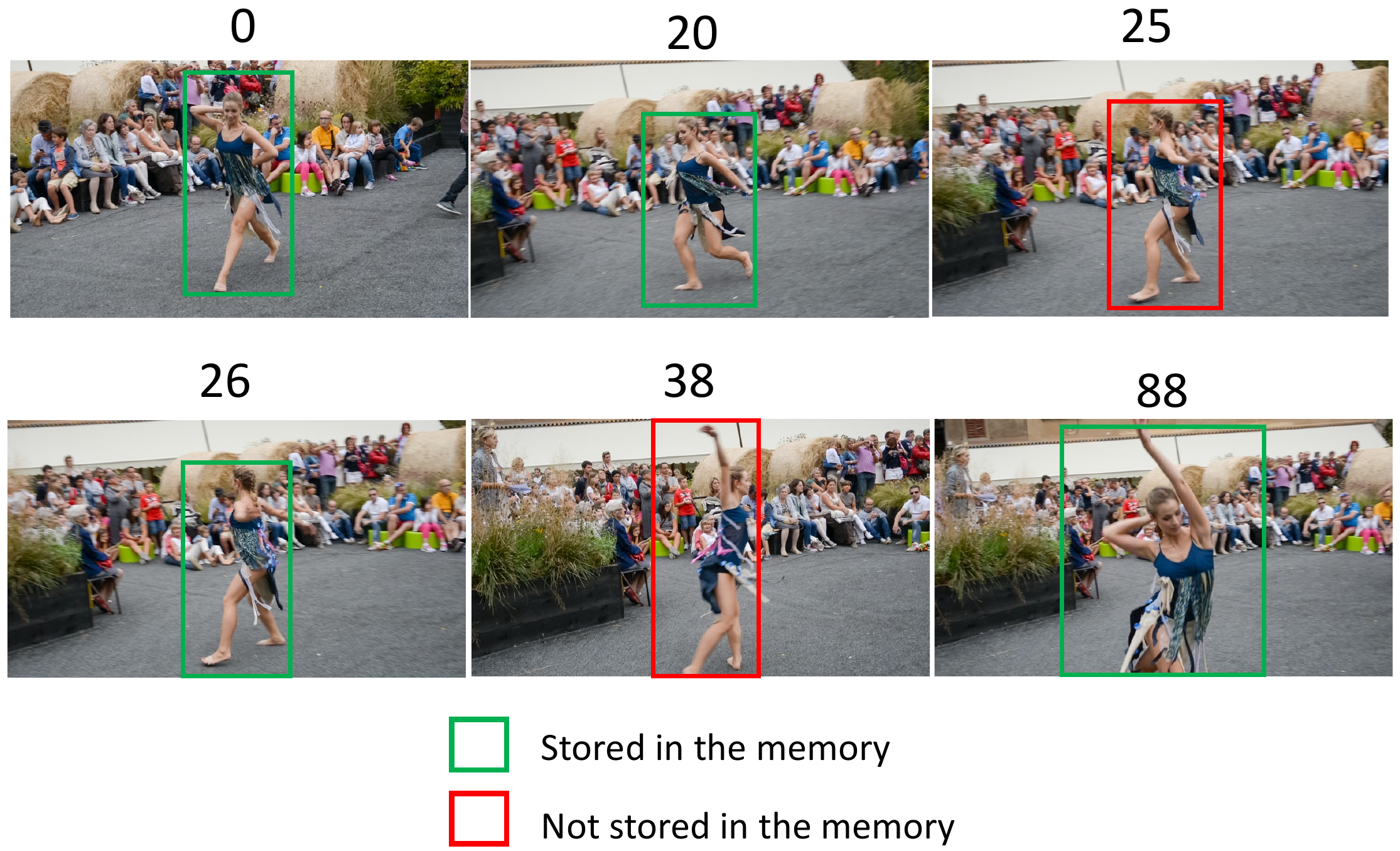}
\end{center}
\caption{After the memory frame 20, the person turns around. The same happens for the memory frame 26. These events make the target appearance change significantly. It can be seen that the memory learned to store useful frames that have different appearances}
\label{fig: mem_dance_twirl}
\end{figure*}

In this section, we show what is stored in the memory. In Fig.~\ref{fig: mem_dance_twirl}, the frames stored in the memory usually show significant appearance changes. Please refer the the caption for more details.

\section{Video Prediction - Qualitative Results}
\begin{figure*}[ht]
\begin{center}
\includegraphics[width=1.0\linewidth]{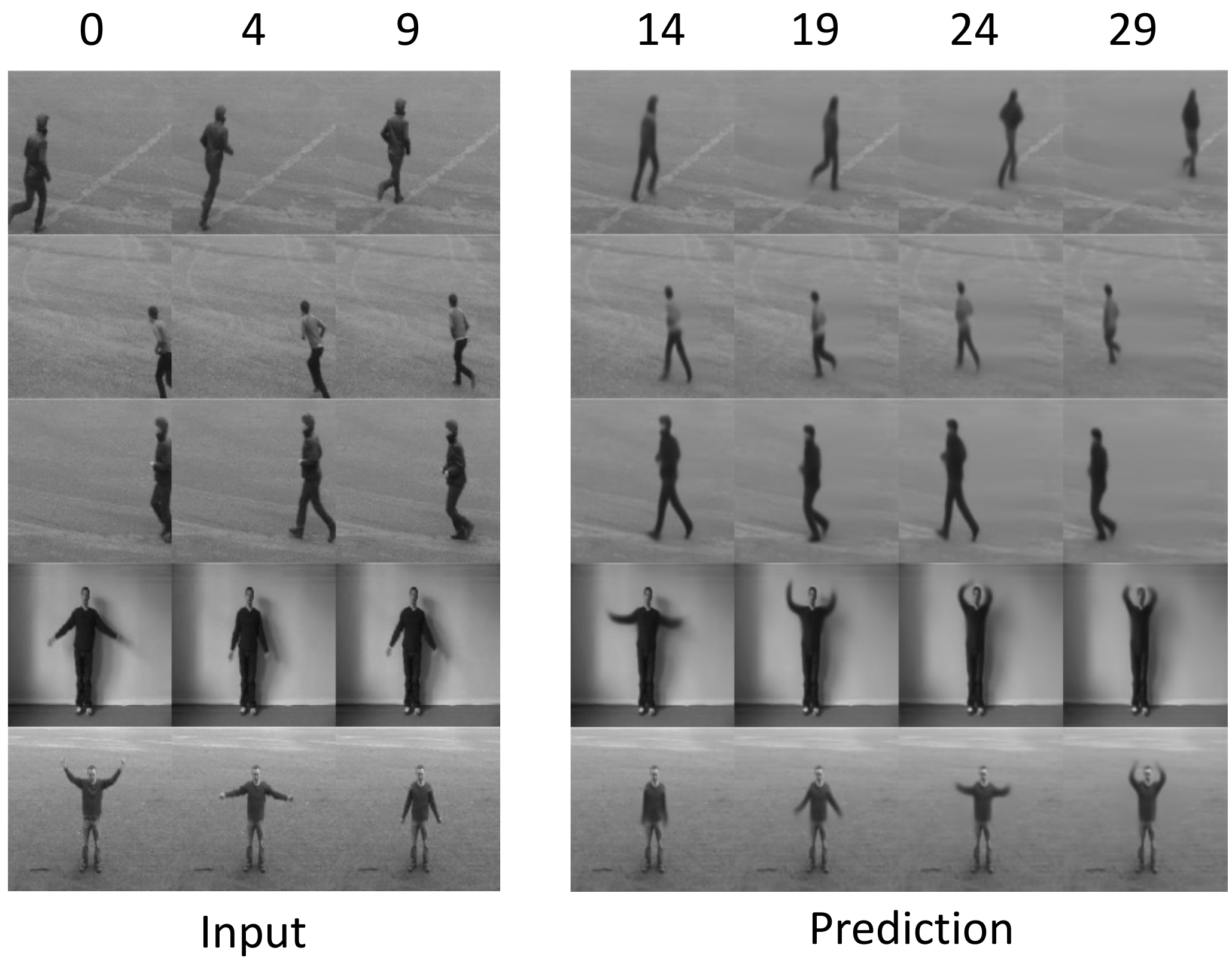}
\end{center}
\caption{Qualitative results for video prediction.}
\label{fig: video_pred_good}
\end{figure*}

\begin{figure*}[ht]

\begin{center}
\includegraphics[width=1.0\linewidth]{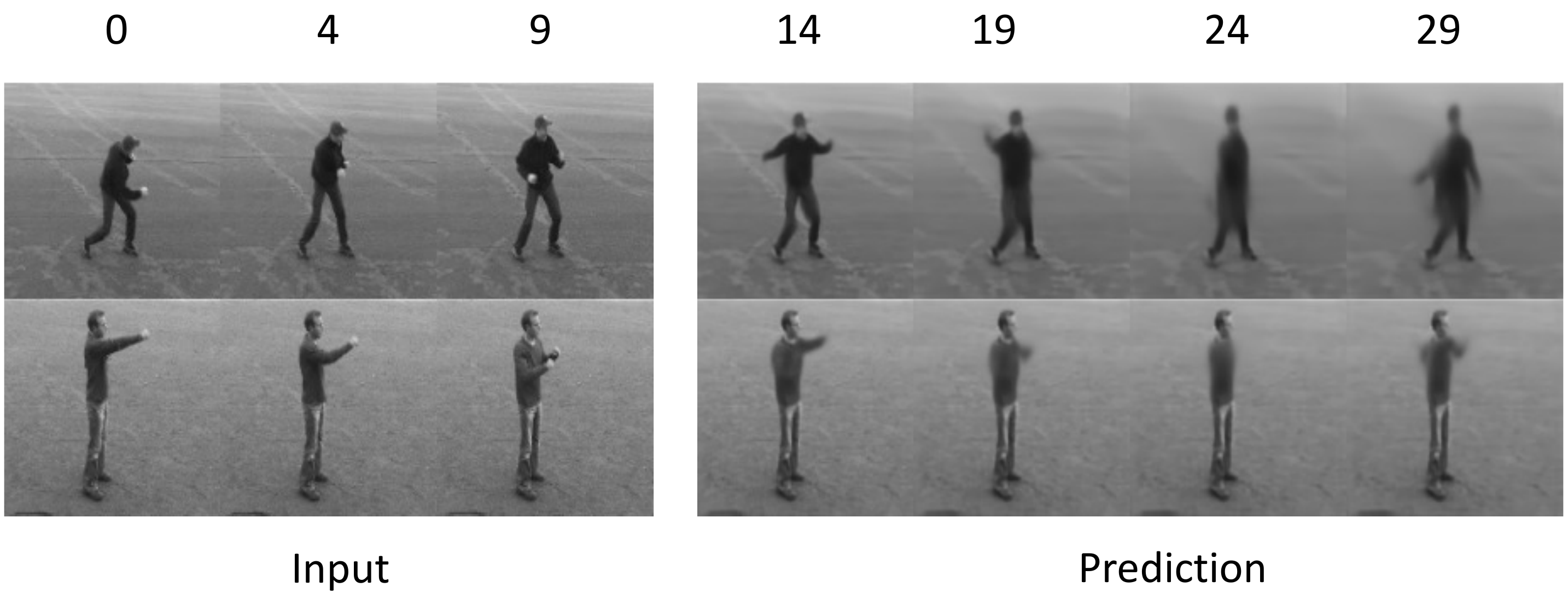}
\end{center}
\caption{Qualitative results for video prediction - Failure cases.}
\label{fig: video_pred_bad}

\end{figure*}

We show more qualitative results in Fig.~\ref{fig: video_pred_good}. In Fig.~\ref{fig: video_pred_bad}, we show some failure cases of our model. In these 2 videos, our model fails to predict the correct position of the arms probably because the arms are moving fast.

\section{Video Object Segmentation - Qualitative Results.}
\begin{figure*}[ht]
\begin{center}
\includegraphics[width=1.0\linewidth]{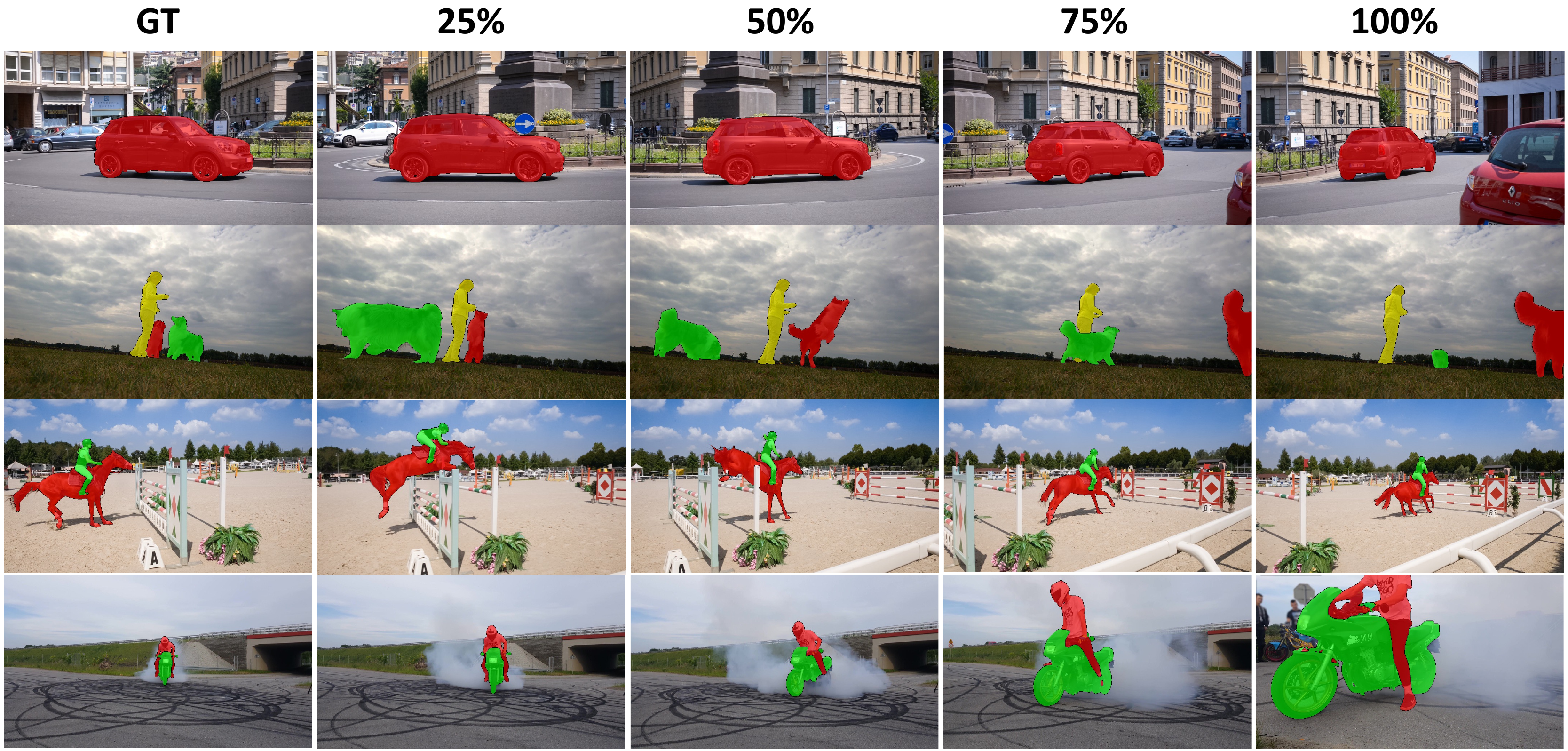}
\end{center}
\caption{Qualitative results for VOS.}
\label{fig: vos_good}
\end{figure*}

\begin{figure*}[ht]
\begin{center}
\includegraphics[width=1.0\linewidth]{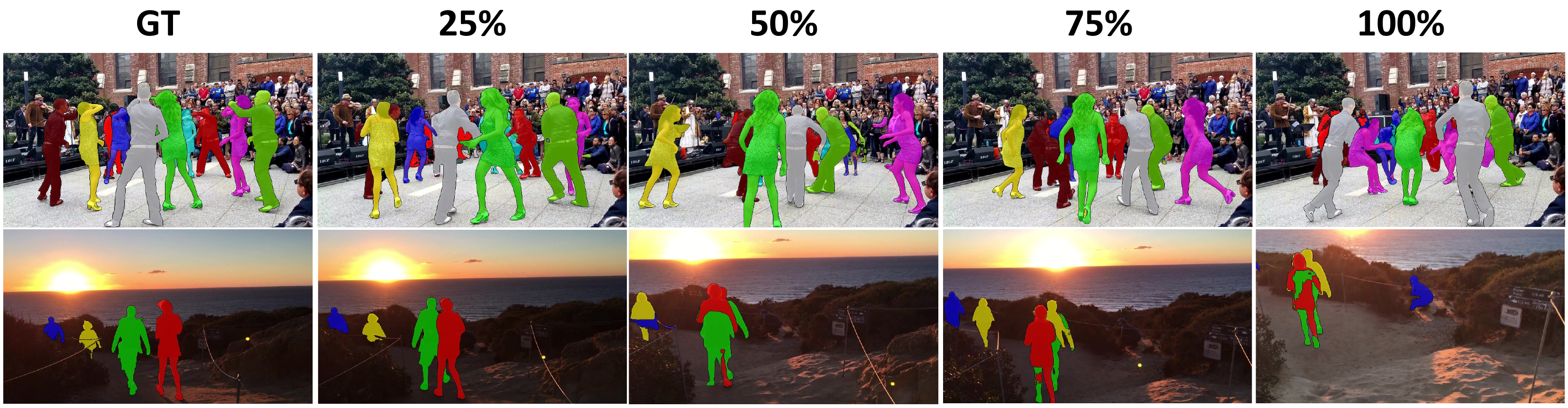}
\end{center}
\caption{Qualitative results for VOS - Failure cases.}
\label{fig: vos_bad}
\end{figure*}

We show more qualitative results for the VOS task in Fig.~\ref{fig: vos_good} and failure cases in Fig.~\ref{fig: vos_bad}. Fig.~\ref{fig: vos_bad} shows that our method sometimes assigns the same object ID to different objects when there are multiple objects of the same types in the scene.

\end{document}